# Algorithms for Learning Decomposable Models and Chordal Graphs

Luis M. de Campos          Juan F. Huete
Departamento de Ciencias de la Computación e I.A.
E.T.S.I. Informática, Universidad de Granada
18071 - Granada, SPAIN
(lci@decsai.ugr.es, jhg@decsai.ugr.es)

## Abstract

Decomposable dependency models and their graphical counterparts, i.e., chordal graphs, possess a number of interesting and useful properties. On the basis of two characterizations of decomposable models in terms of independence relationships, we develop an exact algorithm for recovering the chordal graphical representation of any given decomposable model. We also propose an algorithm for learning chordal approximations of dependency models isomorphic to general undirected graphs.

## 1 INTRODUCTION

Graphical models are knowledge representation tools used by an increasing number of researchers, particularly from the Uncertainty in Artificial Intelligence community. The reason for the success of graphical models is their capacity to represent and handle independence relationships (which have proved crucial for the efficient management and storage of information), as well as uncertain information.

Among the different kinds of graphical models, we are particularly interested in undirected and directed graphs (which, in a probabilistic context, are usually called Markov networks and Bayesian networks, respectively). Each one has its own merits and shortcomings, but neither of these two representations has more expressive power than the other: there are independence relationships that can be represented by means of directed graphs and cannot be represented by using undirected ones, and reciprocally. However, there is a class of models that can be represented by means of both directed and undirected graphs, which is precisely the class of decomposable models (Pearl 1988). Decomposable models also possess important properties, relative to factorization and parameter estimation, which make them quite useful. So, these models have been studied and characterized in many different ways (Beeri et al. 1983, Lauritzen 1989, Lauritzen et al. 1984, Pearl 1988, Whittaker 1991). For example, decomposable models have been characterized as the kind of dependency models isomorphic to chordal graphs (Lauritzen et al. 1984, Whittaker 1991).

Chordal graphs are important for graphical modeling, because local updating of probabilities in graphical models is based on a previous transformation of the initial graph structure into a chordal graph (Lauritzen and Spiegelhalter 1988, Pearl 1988). So, from the perspective of learning models from data, it may be interesting to estimate directly the chordal graph from the available data, instead of first learning the initial graph and after converting it into a chordal graph. The objective of this work is precisely to develop algorithms for learning chordal graphs from data. Our algorithms belong to the kind of learning methods which obtain the graph structure by testing conditional independence relationships among variables, and they are based on a previous work where the independence properties that characterize decomposable models were identified (de Campos 1996).

The rest of the paper is organized as follows. In Section 2 we describe several concepts which are basic for subsequent development. We also briefly review decomposable models and their representation using chordal graphs, as well as two characterizations of decomposable models in terms of independence relationships, which constitute the basis of our algorithms. Section 3 presents an algorithm for recovering the chordal graph representing any given decomposable model. In Section 4, we develop another algorithm (which is an extension of the previous one) for learning graphs which are minimal chordal approximations on dependency models isomorphic to undirected graphs. Finally, Section 5 contains the concluding remarks and some proposals for future work.

## 2 PRELIMINARIES

A *Dependency Model* (Pearl 1988) is a pair $M = (U, I)$, where $U$ is a finite set of elements or variables, and $I(.,.|.)$ is a rule that assigns truth values to a three place predicate whose arguments are disjoint subsets of $U$. Single elements of $U$ will be denoted by standard or Greek lowercase letters, whereas subsets of $U$ will be represented by capital letters. The interpretation of the conditional independence assertion $I(X, Y|Z)$ is that having observed $Z$, no additional information about $X$ could be obtained by also observing $Y$. For example, in a probabilistic model, $I(X, Y|Z)$ holds if and only if

$$P(\mathbf{x}|\mathbf{z}, \mathbf{y}) = P(\mathbf{x}|\mathbf{z}) \text{ whenever } P(\mathbf{z}, \mathbf{y}) > 0,$$

for every instantiation $\mathbf{x}$, $\mathbf{y}$ and $\mathbf{z}$ of the sets of variables $X$, $Y$ and $Z$.

A graphical representation of a dependency model $M = (U, I)$ is a direct correspondence between the elements in $U$ and the set of nodes in a given graph, $G$, such that the topology of $G$ reflects some properties of $I$. The topological property selected to represent independence assertions depends on the type of graph we use: *separation* for undirected graphs and *d-separation* (Pearl 1988) for directed acyclic graphs (dags), both denoted by $\langle X, Y|Z \rangle_G$.

Given a dependency model, $M$, an undirected graph (a dag, respectively), $G$, is said to be an *I-map* if every separation (d-separation, respectively) in $G$ implies an independence in $M$: $\langle X, Y|Z \rangle_G \Rightarrow I(X, Y|Z)$. On the other hand, an undirected graph (a dag, resp.), $G$, is called a *D-map* if every independence relation in the model implies a separation (d-separation resp.) in the graph: $I(X, Y|Z) \Rightarrow \langle X, Y|Z \rangle_G$. A graph, $G$, is a *Perfect map* of $M$ if it is both an I-map and a D-map. $M$ is said to be *graph-isomorphic* if a graph exists which is a perfect map of $M$.

The class of dependency models isomorphic to undirected graphs has been completely characterized (Pearl and Paz 1985) in terms of five properties or axioms satisfied by the independence relationships within the model:

(C1) Symmetry:

$(I(X, Y|Z) \Rightarrow I(Y, X|Z)) \; \forall X, Y, Z \subseteq U.$

(C2) Decomposition:

$(I(X, Y \cup W|Z) \Rightarrow I(X, Y|Z)) \; \forall X, Y, W, Z \subseteq U.$

(C3) Strong Union:

$(I(X, Y|Z) \Rightarrow I(X, Y|Z \cup W)) \; \forall X, Y, W, Z \subseteq U.$

(C4) Intersection:

$(I(X, Y|Z \cup W) \text{ and } I(X, W|Z \cup Y) \Rightarrow I(X, Y \cup W|Z)) \; \forall X, Y, W, Z \subseteq U.$

(C5) Transitivity:

$(I(X, Y|Z) \Rightarrow I(X, \gamma|Z) \text{ or } I(\gamma, Y|Z) \; \forall \gamma \in U \setminus (X \cup Y \cup Z)) \; \forall X, Y, Z \subseteq U.$

**Theorem 1 (Pearl and Paz, 1985)** *A dependency model $M$ is isomorphic to an undirected graph if, and only if, it satisfies the axioms C1–C5.*

The graph associated with the dependency model $M$, such that conditional independence in $M$ is equivalent to separation in this graph, is $G_M = (U, E_M)$, where the set of edges $E_M$ is

$$E_M = \{\alpha\text{–}\beta \mid \alpha, \beta \in U, \neg I(\alpha, \beta|U \setminus \{\alpha, \beta\})\}.$$

On the other hand, the class of dependency models isomorphic to dags is considerably more difficult to characterize. It has been suggested (Pearl 1988) that the number of axioms required for a complete characterization of the d-separation in dags is probably unbounded.

Graphical models are not only convenient means of expressing conditional independence statements in a given domain of knowledge, they also convey information necessary for decisions and inference, in the form of numerical parameters (probabilities) quantifying the strength of each edge. The assignment of parameters to a graphical model is also quite different for undirected and directed graphs. In the case of directed acyclic graphs, this is a simple matter: we only have to assign to each variable $x_i$ in the dag a conditional probability distribution for every instantiation of the variables that form the parent set of $x_i$, $\pi(x_i)$. The product of these local distributions constitutes a complete and consistent specification, i.e., a joint probability distribution (which also preserves the independence relationships displayed by the dag). The case of undirected graphs is different: constructing a complete and consistent quantitative specification while preserving the dependence structure of an arbitrary undirected graph can be done using the method of Gibb's potentials (which assigns compatibility functions to the cliques of the graph), but it is considerably more complicated, in terms of both computational effort and meaningfulness of the parameters, than the simple method used for dags.

### 2.1 DECOMPOSABLE MODELS AND CHORDAL GRAPHS

Some dependency models representable by means of a special class of undirected graphs do not present the previous quantification problem. These are the so called decomposable models, which also exhibit a number of important and useful additional properties. The most appropriate way of defining decomposable models to our interests, which mainly lie in graphical modeling, is based on a graph-theoretic concept: chordal graphs.

**Definition 1** *An undirected graph is said to be chordal if every cycle of length four or more has a chord, i.e., an edge linking two non-adjacent nodes in the cycle.*

**Definition 2** *A dependency model is decomposable if it is isomorphic to a chordal graph.*

One important property satisfied by every chordal graph $G$, which in fact characterizes chordal graphs (Beeri et al. 1983), is that the edges of $G$ can be directed acyclically so that every pair of converging arrows emanates from two adjacent nodes. From this property, it can be deduced (Pearl 1988) that the class of dependency models that may be represented by both a dag and an undirected graph is precisely the class of decomposable models.

Another crucial property of chordal graphs is that their *cliques* (i.e., the largest subgraphs whose nodes are all adjacent to each other) can be joined to form a tree $T$, called the *join tree*, such that any two cliques containing a node $\alpha$ are either adjacent in $T$ or connected by a chain of $T$ made entirely of cliques that contain $\alpha$ (Beeri et al. 1983).

This result has important consequences for probabilistic modeling: the joint probability distribution factorises into the product of marginal distributions on cliques (Lauritzen et al. 1984, Pearl 1988, Whittaker 1991); moreover, maximum likelihood estimates of the model are directly calculable (Whittaker 1991). As a consequence the compatibility functions used to quantitatively specify the model, have a clear meaning and can be easily estimated. Additionally, the tree structure of the cliques in a chordal graph facilitates recursive updating of probabilities. In fact, one of the most important algorithms for propagation (i.e., updating using local computations) of probabilities in undirected graphs and dags, is based on a transformation of the given undirected graph (dag, respectively) into a chordal graph, by triangulating (moralizing and next triangulating, respectively) the graph (Lauritzen and Spiegelhalter 1988).

## 2.2 CHARACTERIZATIONS OF DECOMPOSABLE MODELS

Recently, two characterizations of decomposable models (or equivalently, of chordal graphs) have been established (de Campos 1996). They are based on identifying the set of properties or axioms that a collection of independence relationships must satisfy, in order to be representable by a chordal graph.

Let us consider the following two axioms:

(C6) Strong Chordality:

$(I(\alpha, \beta | Z \cup \gamma \cup \delta)$ and $I(\gamma, \delta | U \setminus \{\gamma, \delta\}) \Rightarrow I(\alpha, \beta | Z \cup \gamma)$ or $I(\alpha, \beta | Z \cup \delta))$ $\forall \alpha, \beta, \gamma, \delta \in U$ $\forall Z \subseteq U \setminus \{\alpha, \beta, \gamma, \delta\}$.

(C8) Clique-separability:

$(I(\alpha, \beta | U \setminus \{\alpha, \beta\}) \Rightarrow \exists W \subseteq U \setminus \{\alpha, \beta\}$ such that $I(\alpha, \beta | W)$ and either $|W| \leq 1$ or $\neg I(\gamma, \delta | U \setminus \{\gamma, \delta\}) \forall \gamma, \delta \in W)$ $\forall \alpha, \beta \in U$.

Axiom C8 asserts that whenever two nodes $\alpha$ and $\beta$ are not adjacent (are independent), we can find a separating set whose nodes are all adjacent to each other, i.e., a complete separating set. Axiom C6 establishes a condition that allows us to reduce the size of the conditioning set separating two variables $\alpha$ and $\beta$, namely that two of the variables in this set are conditionally independent. Equivalently, C6 says that if a separator of $\alpha$ and $\beta$ is not complete, then it has a proper subset which is still a separator of $\alpha$ and $\beta$; moreover, we can find this subset by removing, from the initial separator, one of the nodes causing its incompleteness.

**Theorem 2 (de Campos, 1996)** *A dependency model $M$ is isomorphic to a chordal graph if, and only if, it satisfies the axioms C1–C5 and either C6 or C8.*

## 3 RECOVERING CHORDAL GRAPHS FROM DECOMPOSABLE MODELS

In this section we develop an algorithm for learning the chordal graph corresponding to any given decomposable dependency model. There are, basically, two general approaches to the problem of learning graphical models: methods based on conditional independence tests, and methods based on a scoring metric. The algorithms based on independence tests carry out a qualitative study of the dependence and independence relationships among the variables in the domain (obtained, for example, from a database by means of conditional independence tests), and try to find a network representing these relationships as far as possible. The main computational cost of this class of algorithms is due to the number and the complexity of the independence tests. Our algorithm belongs to this class.

As the problem of learning graphical models from data (either using independence tests or scoring metrics) is computationally very complex, some algorithms (Spirtes et al. 1993) start from a complete undirected graph, and then try to remove edges by testing for conditional independence between the linked nodes, but using conditioning sets as small as possible (thus reducing the complexity and increasing reliability). Our algorithm adopts this methodology too, but it also takes into account the previous axiomatic characterizations of decomposable models to further reduce the number and complexity of the tests. Indeed, the basic independence properties of decomposable models, C6 and C8, could guide us in the design of more efficient algorithms for learning chordal graphs:

- If we rewrite the property C6 in the following way:
  $\neg I(\alpha,\beta|Z\cup\gamma)$ and $\neg I(\alpha,\beta|Z\cup\delta)$ and $I(\alpha,\beta|Z\cup\gamma\cup\delta) \Rightarrow \neg I(\gamma,\delta|U\setminus\{\gamma,\delta\})$,
  
  then we could use it as a rule that simultaneously allows us to remove the edge $\alpha$–$\beta$ from the current graph, and to fix the edge $\gamma$–$\delta$ as a true edge in the graph.

- Similarly, the property C8 could give rise to the following rule: if we are trying to remove an edge $\alpha$–$\beta$ from the current graph, by testing conditional independence statements like $I(\alpha,\beta|W)$, then discard as candidate separating sets those sets $W$ whose nodes are not all adjacent to each other.

All these ideas give rise to the algorithm displayed in Figure 1.

First, the algorithm removes edges by testing conditional independence relationships of order zero and one (lines 3 to 15). Next, it considers, iteratively, conditional independence relationships or order 2, 3,... At this stage, the algorithm removes edges but it also can fix edges by applying the rules derived from axioms C6 and C8 (thus reducing the number of necessary tests). At any step, the algorithm only needs to consider, as candidate conditioning sets to separate $x$ and $y$, those subsets of either the current set of nodes adjacent to $x$ or to $y$ ($\text{Adj}_G(x)$ or $\text{Adj}_G(y)$).

The following theorem proves the correctness of the algorithm.

**Theorem 3** *If a dependency model $M$ is decomposable, then the graph $G$ obtained by the algorithm in Figure 1 is a chordal graph isomorphic to $M$.*

**Proof.** 1.- The algorithm only removes edges when it finds an independence relationship (lines 4, 12 and 36), so that all these edges are eliminated correctly. Therefore, if $G_0$ is the true chordal graph isomorphic to $M$, and $G$ is the graph obtained at any step by the algorithm, we have $G_0 \subseteq G$, or, in other words, the algorithm never removes an edge from $G_0$.

2.- On the other hand, we are going to prove that if $x$ and $y$ are not adjacent nodes in $G_0$, then we can find a minimum separator of $x$ and $y$ contained in $S_x$ (analogous for $S_y$). We know that in any undirected graph $G$, two non adjacent nodes $x$ and $y$ are always separated by the set of nodes adjacent to $x$ and by the set of nodes adjacent to $y$ ($\langle x,y|\text{Adj}_G(x)\rangle_G$ and $\langle x,y|\text{Adj}_G(y)\rangle_G$). Moreover, as only the edges which are not in $G_0$ can be eliminated, we have that, at every step of the algorithm, $\text{Adj}_{G_0}(x) \subseteq \text{Adj}_G(x) \setminus \{y\}$, for any two nodes $x$ and $y$ non adjacent in $G_0$ (and $\text{Adj}_{G_0}(y) \subseteq \text{Adj}_G(y) \setminus \{x\}$). We will also use the fact that in any undirected graph there are exactly $m$ disjoint chains linking two non adjacent nodes $x$ and $y$ if and only if the size of any minimum set separating $x$ and $y$ is equal to $m$ (i.e., $I(x,y|Z)$, $|Z| = m$ and $\neg I(x,y|W)$ $\forall W$ s.t. $|W| < m$). Let $x$ and $y$ be two non adjacent nodes in $G_0$ and let $m$ be the size of any minimum set separating $x$ and $y$; we shall see that the algorithm finds a minimum separating set (of size $m$) and removes the edge $x$–$y$. It is suffice to show that when $n = m$ both sets $S_x$ and $S_y$ will contain a separator subset: Let $Z$ be any minimum separating set of $x$ and $y$, $Z = \{z_1,\ldots,z_m\}$. From C6 we can deduce that $Z$ is a complete set. Each node $z_i$ blocks one of the $m$ disjoint chains, $c_i$, linking $x$ and $y$. Let us consider any node $z_i$ from $Z$: if $z_i \in \text{Adj}_{G_0}(x)$ then $z_i \in \text{Adj}_G(x) = S_x$. Suppose that $z_i \notin \text{Adj}_{G_0}(x)$; there are at least $m$ disjoint chains linking $x$ and $z_i$: the subchain of $c_i$ going from $x$ to $z_i$, and the $m - 1$ subchains of $c_j$, $j \neq i$, going from $x$ to $z_j$ plus the edge $z_j$–$z_i$. If there are more than $m$ disjoint chains linking $x$ and $z_i$, then $x$ and $z_i$ cannot be separated by a set of size $m$, hence at this stage of the algorithm $z_i$ still belongs to $S_x$. If there are not more disjoint chains from $x$ to $z_i$, then we can replace $z_i$ by another node $s_i$ in the chain $c_i$ which belongs to $S_x$, and the set $(Z \setminus \{z_i\}) \cup \{s_i\}$ is still a minimum separator of $x$ and $y$. By using this reasoning for all the nodes in $Z$ we can find a minimum separator of $x$ and $y$ contained in $S_x$.

3.- The algorithm only turns an edge $x$–$y$ as permanent in some of the following situations:

-Because it has explored all the subsets of the set of adjacent nodes (condition $|S| < n$, line 23): in this case $x$–$y$ has to be a true edge of $G_0$.

-By applying the axioms C6 (lines 34-38; note that the separator set $N$ is necessarily of minimum size) or C8 (lines 24-27), which are true properties for chordal graphs; therefore, the edge $x$–$y$ has to be also a true edge of $G_0$.

In conclusion, the algorithm never turns an edge which is not in $G_0$ as permanent, and removes all the non-permanent edges which are not in $G_0$, so that $G \subseteq G_0$ when the algorithm finishes. Therefore, $G_0 = G$. □

```
1  Let G(U) be a complete undirected graph.
.  mark all the edges in G as non-permanent.
.  For every pair of nodes x,y ∈ U do
.      If I(x,y|∅) remove the edge x-y from G.
5  For every x,y adjacent in G do {
.    V = U \ {x,y}.
.    separated = False.
.    While (V ≠ ∅ and !separated) do {
.      select a node z ∈ V.
10     V = V \ {z}.
.      If I(x,y|z) {
.        remove the edge x-y from G.
.        separated = True }
.    }
15 }
.  n = 2.
.  Repeat {
.   For every x,y adjacent in G
.   s.t. the edge x-y is not permanent do {
20    S_x = Adj_G(x) \ {y}.
.     S_y = Adj_G(y) \ {x}.
.     If ( |S_x| ≤ |S_y| ) S = S_x else S = S_y.
.     If ( |S| < n ) mark x-y as permanent.
.     Else If (there is not any subset N ⊆ S,
25         |N| = n s.t. the nodes in N are all
.          adjacent among each other)
.       mark the edge x-y as permanent.
.     Else {
.       separated = False.
30      While (there are subsets N ⊆ S, |N| = n
.        s.t. the nodes in N are all adjacent
.        among each other and !separated) do {
.          select one of these subsets, N.
.          If I(x,y|N) {
35           separated = True.
.            remove the edge x-y from G.
.            mark all the edges joining the
.            nodes in N as permanent. }
.       }
40    }
.   } /* For */
.   n = n + 1.
.  }Until all the edges in G are permanent.
```

Figure 1: Algorithm for Learning Chordal Graphs from Decomposable Models

```
For every u,v adjacent in G which are not
in N, such that the edge u-v is not
permanent do
    If I(u,v|N) remove the edge u-v.
```

Figure 2: Additional Steps Necessary for Learning Minimal Chordal I-maps

## 4  LEARNING CHORDAL I-MAPS FROM UNDIRECTED GRAPHS

As we have already commented, practical use of graphical models requires that the undirected graph or the dag representing the model be converted into a chordal graph. So, it may be more efficient to construct directly the chordal graph from the available data, instead of first estimating the graph and next transforming it into a chordal graph. In this section, we consider the previous problem for the case of undirected graphs (more precisely, for dependency models isomorphic to undirected graphs).

So, we start from a dependency model $M$ isomorphic to an undirected (hidden) graph $G_0$, which may be non chordal (i.e., $M$ verifies the properties C1–C5 but not necessarily C6 or C8). Our objective is to directly learn an appropriate chordal representation of $M$. Obviously, if $G_0$ is not chordal, no chordal graph $G$ can be isomorphic to $M$, so that we only require that the chordal graph to be constructed is an I-map of $M$.

If we apply the algorithm in the previous section to this situation, we cannot guarantee that the resultant graph is chordal (the algorithm could obtain a chordal graph, but this depends greatly on the order in which the independence tests are performed). The problem arises because the algorithm, once an edge $u$–$v$ has been eliminated by finding a true independence statement $I(u,v|W)$ (and the edges linking nodes in $W$ have been fixed), does not take into account that other edges could also be removed by using the same separating set $W$. This fact gives rise to the possibility of fixing, at subsequent steps, more edges than necessary. For example, if after removing the edge $u$–$v$, the algorithm finds $I(r,s|Z)$, then it will eliminate the edge $r$–$s$ and will fix all the edges linking nodes in $Z$; in case that $I(r,s|W)$ were also true, we could have removed the edge $r$–$s$ without need of fixing additional edges from $Z$. This addition of unnecessary edges may create cycles without chords in the graph, thus preventing it of being chordal.

However, a simple modification of the basic algorithm, that essentially controls more the order in which the independence tests are carried out, allows us to solve the problem: every time the algorithm is able of removing an edge $u$–$v$ by using a separating set $W$, it will also try to eliminate other edges using the same candidate separating set $W$. The additional steps to be inserted to the algorithm in Figure 1 (between lines 36 and 37), which guarantee that the resultant modified algorithm will find a minimal chordal I-map, are displayed in Figure 2.

To show the correctness of the modified algorithm, we need to prove the following previous result [1]:

**Proposition 1** *If the modified algorithm tests an independence relationship $I(x,y|Z)$ and finds it true, then the nodes $x$ and $y$ are separated by $Z$ in $G$, i.e., $\langle x, y|Z\rangle_G$.*

**Proof.** Suppose that $Z = \emptyset$, i.e., $I(x,y|\emptyset)$, and that $\neg\langle x,y|\emptyset\rangle_G$. Then, there is at least a chain in $G$, $xt_1 \ldots t_m y$, linking $x$ and $y$. By using transitivity, from $I(x,y|\emptyset)$ we deduce $I(x,t_1|\emptyset)$ or $I(t_1,y|\emptyset)$. If the first independence were true, then the algorithm would have also found it, and the edge $x$–$t_1$ would have been removed from $G$. If the second independence is true, then by applying once again transitivity, we get $I(t_1,t_2|\emptyset)$ or $I(t_2,y|\emptyset)$. The first independence cannot be true because the algorithm tests it and the edge $t_1$–$t_2$ has not been eliminated from $G$. By repeatedly using the same argument we obtain $I(t_{m-1},t_m|\emptyset)$ or $I(t_m,y|\emptyset)$, and both relationships are in contradiction with the existence of the edges $t_{m-1}$–$t_m$ and $t_m$–$y$ in $G$. Hence we have $\langle x,y|\emptyset\rangle_G$. If $Z = \{z\}$, i.e., $I(x,y|z)$, using the same argument as before (applying transitivity) we can obtain $\langle x,y|z\rangle_G$.

For the case $I(x,y|Z)$, $|Z| \geq 2$, we can prove the result by using an inductive argument: we suppose that the result is true for all the independence tests carried out by the algorithm at previous steps. If $\neg\langle x,y|Z\rangle_G$, then we have in $G$ a chain $xt_1 \ldots t_m y$ linking $x$ and $y$ which does not contain nodes from $Z$, i.e., $t_i \notin Z\ \forall i$. From transitivity we obtain $I(x,t_1|Z)$ or $I(t_1,y|Z)$. If the first relationship is true, then either the algorithm checks it (and then removes the edge $x$–$t_1$, which is not possible), or the algorithm does not check it because the edge $x$–$t_1$ has been fixed at a previous step. But this means that, at a previous step, the algorithm has found the independence $I(u,v|C \cup x \cup t_1)$ and as a consequence it has removed the edge $u$–$v$ and fixed the edge $x$–$t_1$ and perhaps some other edges (the other ways of fixing edges, lines 23 and 27, do not apply in this case: they are only able of fixing true edges of $G_0$, see (de Campos and Huete 1997)). Taking into account the iterative way of working of the algorithm, when it finds an independence relationship, this relationship is necessarily minimal (if an independence using a smaller conditioning set were true, it would have been found previously, see (de Campos and Huete 1997)). So, we can assert $I(u,v|C \cup x \cup t_1)$, $\neg I(u,v|C \cup x)$ and $\neg I(u,v|C \cup t_1)$. These relationships imply that, in the graph $G_0$, there is a chain linking $u$ and $v$ which contains $x$ and no other node from $C \cup t_1$, and there is another chain containing $t_1$ and no other node from $C \cup x$. This also implies that, in $G_0$, there are two different chains $c_u$ and $c_v$ linking $x$ and $t_1$, $c_u$ containing $u$ and $c_v$ containing $v$ (it may also happen that $c_u$ (analogous for $c_v$) does not contain $u$ but there is another chain linking $u$ with some node in $c_u$; this chain does not contain any node from $C \cup x \cup t_1$). As $I(x,t_1|Z)$ is true, these two chains are blocked by $Z$. So, in the chain $c_u$ there is a node $z_1 \in Z$, and in the chain $c_v$ there is another node $z_2 \in Z$. By applying transitivity to $I(u,v|C \cup x \cup t_1)$ twice we can obtain $I(z_1,z_2|C \cup x \cup t_1)$. But, as this test has been checked at a previous step, by the induction hypothesis we can assert $\langle z_1,z_2|C \cup x \cup t_1\rangle_G$; then, the edge $z_1$–$z_2$ would have been removed from $G$ and $Z$ is not a complete separating set, hence the algorithm would not have tested $I(x,y|Z)$, in contradiction with the hypothesis. This argument proves that the relationship $I(x,t_1|Z)$ cannot be true. Now, from $I(t_1,y|Z)$ we deduce $I(t_1,t_2|Z)$ or $I(t_2,y|Z)$. By repeating the same previous reasoning we shall again obtain a contradiction. So, we can assert $\langle x,y|Z\rangle_G$. $\square$

The next theorem proves that the new algorithm works correctly.

**Theorem 4** *If a dependency model $M$ is isomorphic to an undirected graph, then the graph $G$ obtained by the algorithm in Figure 1 enlarged with the steps displayed in Figure 2, is a chordal graph which is an I-map of $M$. Moreover, no other chordal graph included in $G$ is an I-map of $M$.*

**Proof.** The algorithm only removes an edge if it finds an independence relationship. Therefore, no edge of $G_0$ (the true undirected graph isomorphic to $M$) can be eliminated, hence $G_0 \subseteq G$ and $G$ is an I-map of $M$.

Now, let us prove that $G$ is a chordal graph. We shall use the characterization based on the axiom C8. If $\langle x,y|U \setminus \{x,y\}\rangle_G$ is true, then the algorithm has removed the edge $x$–$y$ because it has found an independence relationship $I(x,y|Z)$, thus also fixing all the edges linking the nodes in $Z$ (if $|Z| \geq 2$). Then, using proposition 1, we can be sure that $\langle x,y|Z\rangle_G$ is true. So, we have a set $Z$ such that $\langle x,y|Z\rangle_G$ and either $|Z| \leq 1$ or $\neg\langle z_1,z_2|U \setminus \{z_1,z_2\}\rangle_G$, $\forall z_1, z_2 \in Z$. Therefore, $G$ is chordal.

Finally, let us see that $G$ is a minimal chordal graph which is an I-map of $M$. We shall prove that if any edge $u$–$v$ is removed from $G$ to obtain another graph $G' = G \setminus \{u-v\}$, then either $G'$ is not chordal or is not an I-map of $M$.

If the removed edge is a true edge in $G_0$, we have in this case $\langle u,v|U \setminus \{u,v\}\rangle_{G'}$ but $\neg I(u,v|U \setminus \{u,v\})$ and then $G'$ cannot be an I-map. On the other hand, if the removed edge is not in $G_0$, this means that the algorithm found $I(x,y|Z \cup u \cup v)$ (and this implies $\langle x,y|Z \cup u \cup v\rangle_G$), $\neg I(x,y|Z \cup u)$ and $\neg I(x,y|Z \cup v)$, for some nodes $x$ and $y$. As $G_0 \subseteq G$ and $u$-$v \notin G_0$, then we have $G_0 \subseteq G'$, i.e., $G'$ is also an I-map of $M$. So, from $\neg I(x,y|Z \cup u)$ and $\neg I(x,y|Z \cup v)$ we deduce $\neg\langle x,y|Z \cup u\rangle_{G'}$ and $\neg\langle x,y|Z \cup v\rangle_{G'}$. Finally, from $\langle x,y|Z \cup u \cup v\rangle_G$ and $G' \subseteq G$ we obtain $\langle x,y|Z \cup u \cup v\rangle_{G'}$. Then, we have $\langle x,y|Z \cup u \cup v\rangle_{G'}$, $\neg\langle x,y|Z \cup u\rangle_{G'}$ and $\neg\langle x,y|Z \cup v\rangle_{G'}$ but $\langle u,v|U \setminus \{u,v\}\rangle_{G'}$, and according to the axiom C6, $G'$ cannot be a chordal graph. □

---

[1] Due to space limitations, we do not include detailed versions of the proofs of some results. Complete proofs can be found in (de Campos and Huete 1997)

Therefore, we can guarantee that the output of the algorithm is a minimal chordal I-map of the dependency model $M$. If $M$ is decomposable, the modified algorithm also finds the chordal graph isomorphic to it. Moreover, looking into the algorithm, it can be seen that there are still several independence tests that could be omitted, since their truth values can be known apriori. The key for this reasoning is to see the algorithm from a different perspective: whenever a separating subset $Z$ (verifying $I(x,y|Z)$ for some nodes $x$ and $y$) is found, then, after testing all the independences with the same conditioning set, the graph is split up into a set of complete subgraphs or cliques, $C_1(U_1), \ldots, C_s(U_s)$. For each clique $C_r$, $r = 1, \ldots, s$, its set of nodes $U_r$ can be divided into two disjoint subsets $W_r$ and $Z$, i.e., $U_r = W_r \cup Z$, $W_r \cap Z = \emptyset$, with $|W_r| \geq 1$; moreover, given two different cliques, $C_r$ and $C_t$, the sets $W_r$ and $W_t$ are disjoint too. Furthermore, for any $z_i, z_j \in Z$, the edge $z_i$-$z_j$ has been marked as permanent.

More formally, starting from the true independence statement $I(x,y|Z)$, found by the algorithm, let us define recursively the sets $W_i$ as follows: for $i = 1$, $x_i = x$, $R_i = U \setminus Z$ and $W_i = \{u \in R_i \,|\, \neg I(u, x_i|Z)\}$; for $i \geq 1$, $R_{i+1} = R_i \setminus W_i$ and, if $R_{i+1} \neq \emptyset$, let $x_{i+1} \in R_{i+1}$ and $W_{i+1} = \{u \in R_{i+1} \,|\, \neg I(u, x_{i+1}|Z)\}$. It is clear (using transitivity) that

1. $\neg I(u,v|Z)$ for all $u,v \in W_r$. So, at this step all the nodes in $W_r$ are adjacent among each other. Moreover, as the algorithm never tests the statements $I(u,z|Z)$, $z \in Z$, then $U_r = W_r \cup Z$ is a clique.

2. $I(u,v|Z)$ for all $u \in W_r, v \in W_t, r \neq t$. Therefore, all the sets $W_i$, $i = 1,\ldots,s$, are disjoint among each other.

A similar reasoning can be recursively stated for each clique $C_r$, $r = 1,\ldots,s$. In that case, we only need to look for a subset of variables (the search is reduced to the variables within $U_r$) separating a pair of nodes in $U_r$ connected by a non permanent edge. The splitting process will stop when all the edges in the sub-cliques of $C_r$ are marked as permanent.

Therefore, the following algorithm (see Figure 3) allows us to obtain a minimal chordal I-map of the initial undirected graph $G_0$. The algorithm takes as the input a complete graph $G(U)$ and gives the list of cliques that constitute the different components of the chordal graph as the output.

---

i.- *Initialization*
```
1  Let G^{0-1} be the graph obtained by
   testing independences of order 0 and 1.
.  Let L be the list of cliques in G^{0-1}.
.  Create O, an empty output list.
5  n = 2.
```
ii.- *Searching for the complete substructures*
```
.  Repeat {
.  For each clique C in L do {
.   If |C| <= n+1 {
.    mark all the edges in C as permanent.
10   remove from L the clique C.
.    insert C in O. }
.   Else {
.    div = False
.    While (∃ non permanent edges in C
15    and !div) do {
.     select one of these edges, x - y.
.     While (∃ subsets N ⊂ C \ {x,y} s.t.
.      |N| = n and !div) do {
.      select one of these subsets, N.
20     If I(x,y|N) ) {
.       call Divide(x,y,C,N,L).
.       div = True. }
.     }
.    }
25 }
.  } /* for each*/
.  n = n + 1.
   }Until L be the empty list.
.  Give O as the output.
```
---

Figure 3: Algorithm for Learning Chordal I-maps

In this algorithm, $Divide(x,y,C,N,\mathcal{L})$ is a function that removes from $\mathcal{L}$ the clique $C$ and includes in $\mathcal{L}$ all the sub-cliques that can be obtained from $C$ by checking independence relationships like $I(.,.|N)$. A sub-clique of $C$ will be composed by the variables in $N$ and those variables $u,v \in C$ such that $\neg I(u,v|N)$. This process can be carried out iteratively, as indicated in Figure 4, avoiding some independence tests. Note that, if the test $I(w,z|N)$ is true (line 14) then it can be assured that node $z$ does not belong to the sub-clique $S$ and, therefore, all the tests $I(t,z|N) \,\forall t \in S$ may be omitted. On the other hand, if the independence test $I(w,z|N)$ in line 14 is false, we know that node $z$ will be included in the clique $S$ and, therefore, it is not necessary to check the truth values for the independence relationships $I(t,z|N) \,\forall t \in S$, and $I(t',z|N) \,\forall t' \in S'$, being $S' \neq S$ any other sub-clique in $LC$.

Moreover, since it is possible that there are some edges in the clique $C$ fixed as permanent in a previous step of the algorithm, for instance the edge $x$–$t$, then we can guarantee that the sub-clique including the node $x$, named $SC_x$, will also include the node $t$. So, the sub-clique $SC_x$ is directly initialized including $x$ and all the nodes in $C$ connected with $x$ by a permanent edge. This process is carried out by the $Initialize(SC_x, C)$ procedure.

```
Divide(x, y, C, N, L)
1  mark all the edges in N as permanent.
.  create LC, an empty list of sub-cliques.
.  call to Initialize(SC_x, C).
.  call to Initialize(SC_y, C).
5  insert SC_x, SC_y in LC.
.  mark all the nodes in SC_x, SC_y as visited.
.  While (exists z ∈ C \ N non visited) do {
.    call to Initialize(SC_z, C).
.    mark all the nodes in SC_z as visited.
10   found = False.
.    While (exists S ∈ LC and !found) do {
.      select one of these sub-cliques, S.
.      let w an element in S.
.      If ¬I(w, z|N) {
.15      join SC_z with S.
.        found = True. }
.    }
.    If (!found) insert SC_z in LC.
.  }
.20 remove from L the clique C.
.  For each S ∈ LC do {
.    let S ← S ∪ N a new clique.
.    insert S in L. }
```

Figure 4: Splitting Procedure

Therefore, considering the comments above, we can say that the algorithm in Figures 3 and 4 has reduced to the minimum the number of necessary independence test.

## 5  CONCLUDING REMARKS

We have shown how an axiomatic characterization of decomposable models, in terms of independence relationships, may create desiderata for driving automated construction of chordal graphs from data. We have developed an algorithm for recovering chordal graphs from decomposable models, and another algorithm that finds a minimal chordal I-map of any dependency model isomorphic to an undirected graph. The restriction to graph-isomorphic models is necessary to derive the results. It is still not clear whether our methodology can be adapted to deal with more general models. We have also made the assumption that the data is a perfect representation of the underlying model, i.e., the algorithms can decide in an error-free manner whether specific conditional independence statements are true or false (this amounts to a sample that is infinite in size). This is a problem common to all the learning algorithms based on independence tests. However, our algorithms use conditional independence tests of order as low as possible, thus gaining in efficiency and also in reliability. For future work, we plan to develop algorithms for learning chordal representations of dependency models isomorphic to dags.

### Acknowledgments

This work has been supported by the Spanish Comisión Interministerial de Ciencia y Tecnología (CICYT) under Project n. TIC96-0781.